\documentclass{article} 
\usepackage[authoryear,round]{natbib}
\usepackage{arxiv}
\usepackage{microtype}
\usepackage[colorlinks]{hyperref}
\usepackage{url}
\usepackage{booktabs}
\usepackage{graphicx}
\usepackage{amsmath,amssymb}
\usepackage{enumitem}
\usepackage{xcolor}
\usepackage{geometry}
\usepackage{subcaption}
\usepackage{multirow}
\usepackage{array}
\usepackage{svg}
\usepackage{fnpct}  
\usepackage{cleveref}
\usepackage{tabularx}

\definecolor{LightCyan}{rgb}{0.88,1,1}
\definecolor{darkblue}{rgb}{0, 0, 0.5}
\hypersetup{colorlinks=true, citecolor=darkblue, linkcolor=darkblue, urlcolor=darkblue}

\headerimage{\includegraphics[height=1.8cm]{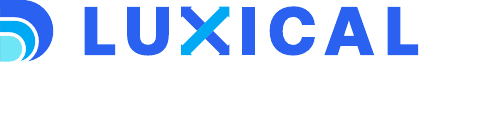}}

\title{
Luxical: High-Speed Lexical-Dense Text Embeddings
}

\author{DatologyAI Team\thanks{See Contributions and Acknowledgments (\Cref{sec:contri}) for full author list.} \\
}
\date{}

\renewcommand{\headeright}{Technical Report}
\renewcommand{\undertitle}{Technical Report}
\renewcommand{\shorttitle}{Luxical: High-Speed Lexical-Dense Text Embeddings}

\begin{document}

\newcommand{\luxicalone}{\texttt{Luxical-One}}

\maketitle
\setcounter{footnote}{0} 

\begin{abstract}
Frontier language model quality increasingly hinges on our ability to organize web-scale text corpora for training. 
Today's dominant tools trade off speed and flexibility: lexical classifiers (e.g., FastText) are fast but limited to producing classification output scores, while the vector-valued outputs of transformer text embedding models flexibly support numerous workflows (e.g., clustering, classification, and retrieval) but are computationally expensive to produce. 
We introduce Luxical, a library for high-speed ``lexical-dense'' text embeddings that aims to recover the best properties of both approaches for web-scale text organization. 
Luxical combines sparse TF--IDF features, a small ReLU network, and a knowledge distillation training regimen to approximate large transformer embedding models at a fraction of their operational cost. 
In this technical report, we describe the Luxical architecture and training objective and evaluate a concrete Luxical model in two disparate applications: a targeted webcrawl document retrieval test and an end-to-end language model data curation task grounded in text classification. 
In these tasks we demonstrate speedups ranging from 3x to 100x over varying-sized neural baselines, and comparable to FastText model inference during the data curation task. 
On these evaluations, the tested Luxical model illustrates favorable compute/quality trade-offs for large-scale text organization, matching the quality of neural baselines. 
Luxical is available as open-source software at \url{https://github.com/datologyai/luxical}.
\end{abstract}

\section{Introduction}
\label{sec:intro}

Large language models (LLMs) have pushed training datasets from billions to trillions of tokens. While webcrawl datasets provide vast quantities of data to meet these demands, much of the information in raw web crawls is redundant, off-distribution, or harmful, while comparatively small subsets of high-value text have demonstrably outsized impact on model capabilities \citep{sorscher2023neuralscalinglawsbeating}. As such, in recent years web-scale corpus organization and quality control have become a necessity for pushing the frontiers of Artificial Intelligence \citep{li2024datacomp}. One promising paradigm for tackling this task is to apply machine learning methods, e.g. supervised classification or representation learning.

A typical tool for ML-based data organization is the use of text embedding transformer models. Though effective for data curation \citep{penedo2024fineweb}, their use often entails a trade-off between representational quality and efficiency. Models designed to score highly on benchmarks such as the Massive Text Embedding Benchmark (MTEB) \citep{muennighoff-etal-2023-mteb} optimize for producing representations suitable to a blend of asymmetrical and symmetrical retrieval tasks and can sacrifice efficiency so aggressively that they become prohibitively expensive to run at web-scale \citep{merrick2024arcticembedscalableefficientaccurate}. In contrast, lexical classification algorithms (which were popular before transformer language models rose to prominence) can operate with extreme efficiency but produce only a single classification score that is unsuitable for certain workloads like clustering and retrieval \citep{joulin2017bag}.

This technical report introduces Luxical, a hybrid lexical--dense embedding modeling methodology that targets a middle ground between lexical classifiers and large transformer text embedding models. Luxical transforms a sequence of tokens into a bag-of-ngrams representation, applies inverse-frequency weighting and normalization to obtain a sparse feature vector, and then transforms this vector into a dense embedding vector using a computationally efficient shallow ReLU network. For training, Luxical models use a Gram-matrix distillation objective that pushes the embedding geometry to match the similarity structure of a larger teacher embedding model, allowing the methodology to leverage advances in large-scale text embedding models without becoming beholden to their slow runtime characteristics. Our goal is not to compete with the strongest transformer encoders on broad benchmarks, but to provide a technically simple point on the speed--quality frontier that is well suited to production data pipelines where dense encoders are too slow and existing lexical methods are not flexible enough.

In this work we describe both the general Luxical methodology and a specific English instantiation, \luxicalone{}, trained via distillation from a strong transformer teacher on English web documents. \luxicalone{} not only serves as a concrete case study of the broader Luxical approach in our experiments, but it is also released as a tool for the broader community.\footnote{The \luxicalone{} model is hosted at \url{https://huggingface.co/DatologyAI/luxical-one}.}

In the remainder of this report we first situate Luxical within the literature on sentence embeddings, web-scale data organization, and lexical retrieval. We then describe the model architecture, tokenization pipeline, and training objective in detail, establish an experimental setup for evaluating \luxicalone{} against lexical and transformer baselines, and discuss how Luxical can be integrated into modern web-scale text-processing workflows.

\section{Motivating Related Work}
\label{sec:background}

Modern transformer-based ``sentence embedding'' models, often derived from BERT and its siamese variants, produce vector-valued representations of natural language applicable to a variety of tasks including clustering, retrieval, and classification \citep{DBLP:journals/corr/abs-1810.04805,reimers-2019-sentence-bert,muennighoff-etal-2023-mteb}. These models set the standard for general-purpose performance, but their compute profiles and memory footprints can be challenging to deploy as the first stage of trillion-token corpus-processing pipelines.

Web-scale corpus construction and filtering methodologies have evolved alongside these embedding models. Early corpora such as the Colossal Clean Crawled Corpus (C4) used heuristic language identification, blocklists, and MinHash-based deduplication to turn Common Crawl into a usable training set for large-scale language modeling \citep{DBLP:journals/corr/abs-1910.10683}. More recent work treats corpus construction partially as a modeling problem. For example, FineWeb and DCLM (DataComp-LM) introduce quality filters based on neural and lexical classifier models, and subsequent corpora show how sensitive downstream performance is to these filters \citep{penedo2024fineweb,li2024datacomp,soldaini2024dolmaopencorpustrillion}. These works motivate the need for efficient embedding backbones that can support model-based filtering and other data organization workflows at web scale.

Lexical and sparse retrieval methods—ranging from TF--IDF and BM25 to more recent models such as SPLADE—exploit term-level structure to achieve strong ranking performance with inverted indexes \citep{robertson2009probabilistic,thibault-2021-splade}. They show that lexical and sparse methods remain competitive and scalable baselines for large retrieval systems, even as dense transformer encoders have become the default choice for many tasks. The continued use of FastText-style models \citep{joulin2017bag,li2024datacomp} similarly illustrates the value of simple bag-of-ngrams architectures in large-scale data pipelines.

\section{Model and Training}\label{sec:model}

\subsection{Lexical--Dense Architecture}

\begin{figure}[htbp]
  \centering
  \includegraphics[width=1.0\textwidth]{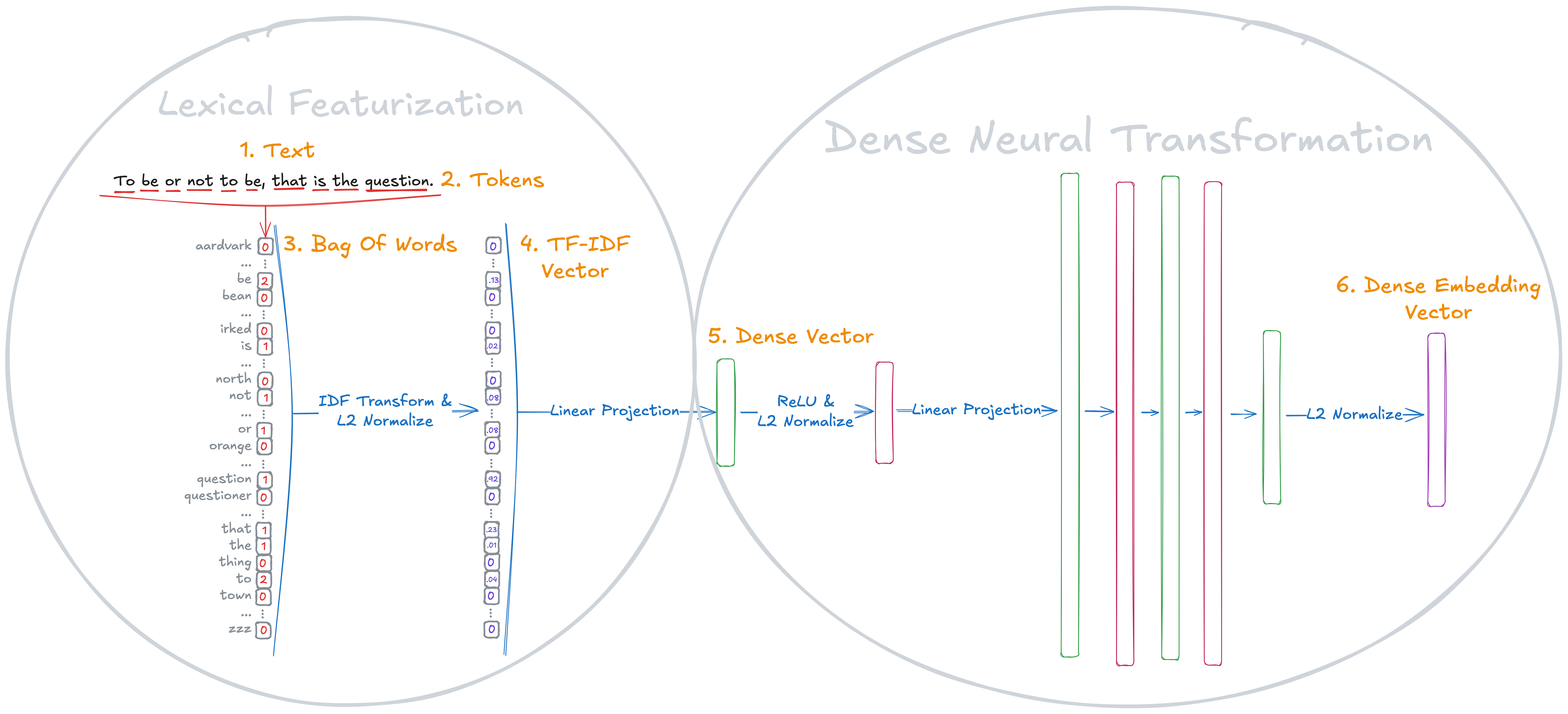}
  \caption{Sparse-to-dense architecture of Luxical. A sparse vector of normalized ngram frequencies (the TF-IDF vector) is projected through a small MLP to produce a dense embedding.}
  \label{fig:sparse-dense}
\end{figure}

Luxical fuses a classical lexical representation into a small ReLU network, yielding dense embeddings while retaining the scalability advantages of bag-of-words methods. Given an input document, Luxical first tokenizes the text and constructs a term-frequency (TF) representation over ngrams from the resulting tokens, similar to the bag-of-ngrams featurization in FastText \citep{joulin2017bag}, but eschewing the hashing trick in favor of exact ngram matching over a predetermined vocabulary of ngrams. This sparse TF vector is then reweighted to account for overall term frequency, and $\ell_2$-normalized to produce a sparse unit vector over the vocabulary. Unlike a purely lexical system, Luxical then applies a small feed-forward ReLU network (which we also refer to as a Multi-Layer Perceptron or MLP for short) to map this sparse term-frequency-inverse-document-frequency (TF--IDF) representation to a dense, normalized embedding. This hybrid design allows the model to encode word-level statistics and simple composition patterns while remaining inexpensive to evaluate.

\subsection{Sparse-to-Dense Projection Efficiency}

The core efficiency property of Luxical models comes from exploiting sparsity in the first linear layer of the MLP. If $\mathbf{s} \in \mathbb{R}^V$ denotes the sparse TF--IDF vector over a vocabulary of size $V$ and $A \in \mathbb{R}^{d \times V}$ denotes the input weight matrix of the MLP, then the matrix--vector product $A \mathbf{s}$ reduces to a sum over the columns corresponding to nonzero entries in $\mathbf{s}$:
\[
  A \mathbf{s} \;=\; \sum_{i \in \mathrm{nz}(\mathbf{s})} s_i \, \mathbf{a}_i,
\]
where $\mathrm{nz}(\mathbf{s})$ indexes nonzero positions and $\mathbf{a}_i$ is the $i$-th column of $A$. In practice, only a small fraction of the vocabulary appears in any given document, so this sparse-by-dense multiplication can be implemented by gathering and scaling a handful of columns. Luxical implements this operation with Numba-optimized kernels \citep{lam2015numba} to achieve high throughput on CPUs. In practice, modern hardware can be so efficient at performing these matrix operations that the tokenization step dominates the overall wall clock time of the Luxical embedding operation, with the ngram counting, sparse-to-dense projection, and extra projections and nonlinearities of the ReLU layers degrading throughput only modestly below that of pure tokenization.

\subsection{Training Objective}\label{subsec:objective}

\begin{figure}[htbp]
  \centering
  \includegraphics[width=1.0\textwidth]{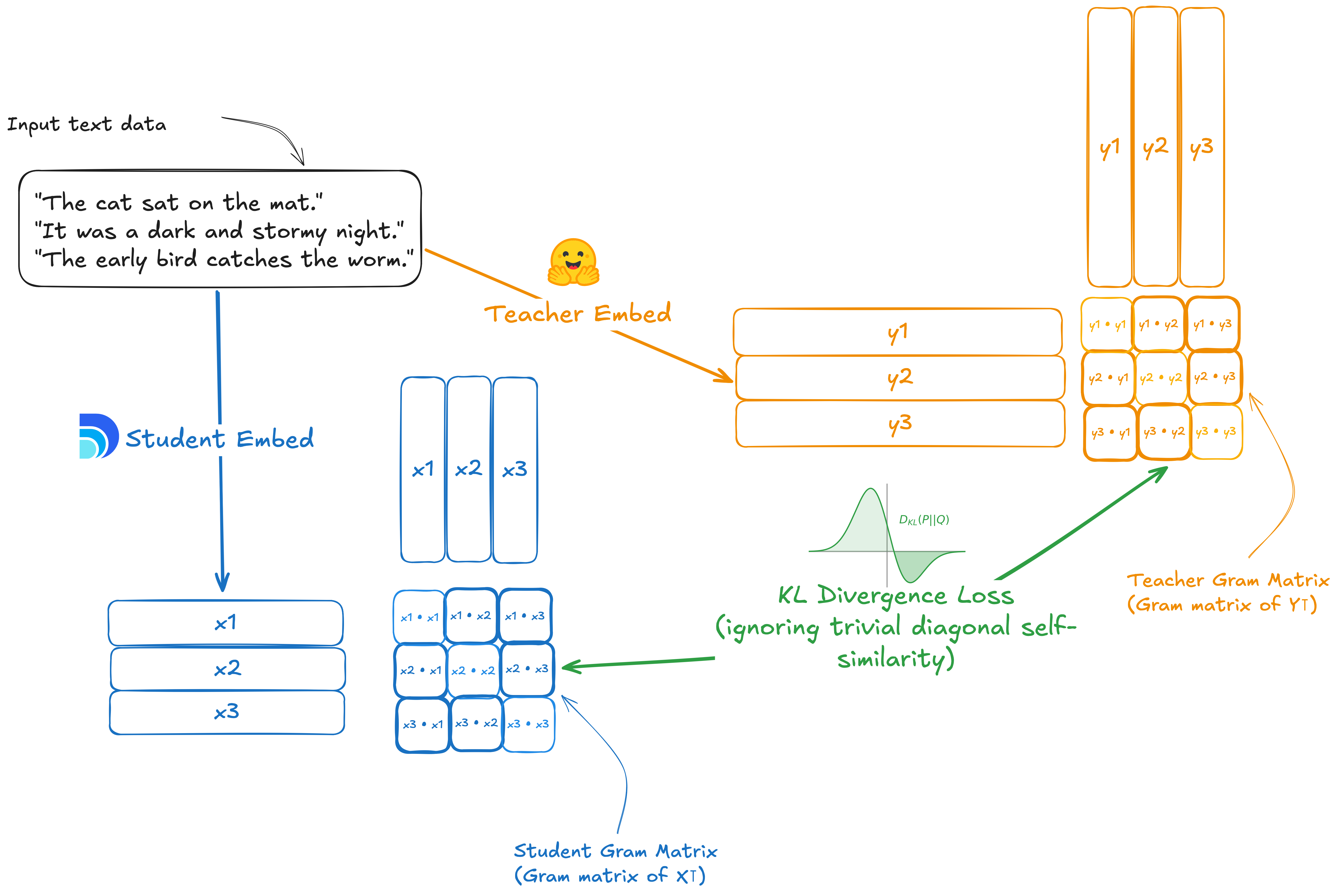}
  \caption{Contrastive distillation loss recommended when training Luxical models.}
  \label{fig:distillation-loss}
\end{figure}

Though the Luxical architecture and codebase are modular and can support a number of training objectives, we choose to implement and study a distillation-style contrastive objective that encourages the Luxical model's embeddings to align the pairwise similarity structure of batches of documents with the structure induced by the embeddings of a teacher model. For a batch of $n$ documents, we compute normalized student embeddings $S \in \mathbb{R}^{n \times d}$ and teacher embeddings $T \in \mathbb{R}^{n \times d_t}$, construct their Gram matrices $G_s = S S^\top$ and $G_t = T T^\top$, and remove the diagonal from both matrices to discard trivial self-similarity (producing $\hat{G}_s$ and $\hat{G}_t$). After temperature scaling (denoting the temperature hyperparameter as $\tau$), we minimize the Kullback--Leibler divergence between the rows of the diagonal-stripped Gram matrices:
\[
  \mathcal{L}_{\mathrm{distill}} \;=\; \tau^2 \cdot \mathrm{KLDiv}\!\bigl(\hat{G}_s / \tau,\;\hat{G}_t / \tau\bigr)
\]
This objective encourages Luxical to replicate the relative similarity pattern induced by the teacher model over each batch, somewhat in the spirit of large-scale weakly supervised contrastive pretraining approaches such as E5 \citep{wang2022text} but leveraging a much stronger supervision signal thanks to the teacher model.

\subsection{Implementation Details}

The implementation is optimized for web-scale deployment on CPU. Though Luxical is written primarily in Python, it relies on a small custom Rust extension for high-throughput tokenization. This \texttt{arrow-tokenize} extension mitigates Python garbage collection overhead by returning tokenized outputs as PyArrow arrays that garbage collect much faster than Python list-of-list outputs. If Luxical were to rely on the built-in Python bindings provided by the \texttt{tokenizers} library, it would have introduced a major performance bottleneck when processing large batches of documents on powerful multi-core CPUs.

Sparse-by-dense projections and IDF scaling are implemented using Numba for efficiency \citep{lam2015numba}. Though IDF scaling weights can be merged into the first layer of the MLP, Luxical performs this as a separate step to keep the parameterization of the layer weights more stable during optimization. Tokenization and embedding computation are pipelined over large batches of documents using the Arrow-based tokenizer to mitigate overheads during high-throughput embedding jobs.

\section{Empirical Evaluation}

We evaluate Luxical through a concrete instantiation of the methodology, the \luxicalone{} model, and a document-level similarity task designed to probe both symmetric semantics and systems-level throughput on FineWeb data.

\subsection{\luxicalone{} Configuration and Training}

\luxicalone{} is an English Luxical model that follows the architecture described in \Cref{sec:model}. The model adopts the BERT uncased tokenizer to segment input text and featurizes documents as TF--IDF-weighted bags over a fixed vocabulary of mined 5-grams. Concretely, we used the Space-Saving Algorithm \citep{spacesavingalgorithm} to identify a vocabulary of approximately two million most frequent 5-grams observed in a sample of the FineWeb corpus as the fixed vocabulary. We also leverage the approximate frequency statistics from this operation to construct a log-scaled IDF scaling vector. The feedforward network of \luxicalone{} maps the 2M-dimensional sparse vector of ngram statistics to dense vectors of sizes 96, 3072, and 3072 with ReLU nonlinearities and $\ell_2$ normalization applied after each projection. The final layer of the network then projects down and $\ell_2$-normalizes once more to produce a compact 192-dimensional embedding vector.

We train \luxicalone{} with the contrastive Gram-matrix distillation objective described in \Cref{subsec:objective}, using teacher embeddings produced by the \texttt{snowflake-arctic-embed-m-v2.0} model \citep{yu2024arcticembed20multilingualretrieval}. We selected this teacher for its relatively modest size (allowing us to embed more documents in the same amount of time compared to other, larger models) and ability to produce small 256-dimensional embedding vectors while retaining high embedding quality (not only did this make data storage and dataloading faster and simpler, it also reduced the cost of computing the teacher Gram matrix during training). We used 50 million English documents sampled from FineWeb as the training corpus and pre-embedded them offline using the teacher model. During training we ran three epochs over this sample, shuffling documents between epochs. Training proceeded with standard mini-batch stochastic optimization (using the Adam optimizer with a warmup-stable-decay learning rate schedule) on CPU only. We used a batch size of 3072, a loss temperature of 3.0, and a peak learning rate of 0.01. We warmed up the learning rate for 5\% of training steps and performed linear learning rate decay to zero learning rate for the final 10\% of training.

\subsection{Throughput Benchmark}\label{subsec:throughput}

\begin{figure}[htp]
  \centering
  \includegraphics[width=\linewidth]{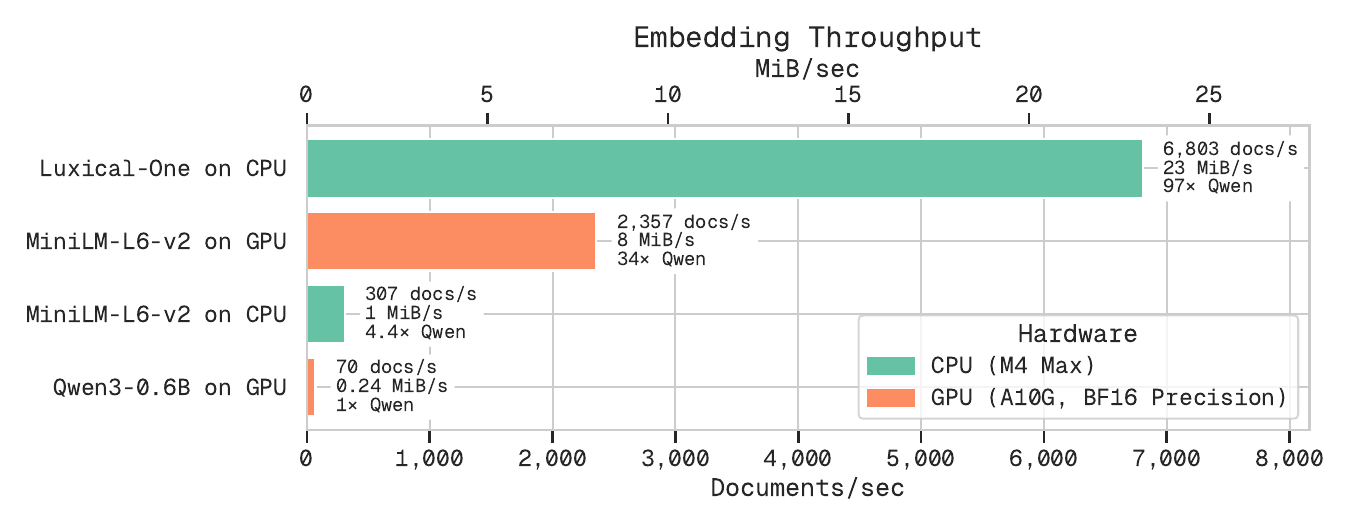}
  \caption{End-to-end throughput (web documents per second) when embedding 100{,}000 FineWeb documents with \luxicalone{} and transformer baselines on an Apple M4 Max CPU and an NVIDIA A10G GPU.}
  \label{fig:throughput}
\end{figure}

To assess how well Luxical delivers high-throughput embedding in practice, we sample 100{,}000 complete FineWeb documents and embed them with \luxicalone{}, \texttt{MiniLM-L6-v2} \citep{sentence_transformers_all_minilm_l6_v2}, and the \texttt{Qwen3-0.6B} embedding model \citep{zhang2025qwen3embeddingadvancingtext}. We report end-to-end throughput in documents per second, including tokenization, under two hardware configurations:
\begin{enumerate}[label=(\roman*),nosep]
  \item an Apple M4 Max laptop CPU, and
  \item an NVIDIA A10G server GPU.
\end{enumerate}
For transformer baselines we evaluate both CPU-only and GPU-accelerated settings where applicable, while \luxicalone{} is evaluated on CPU only, reflecting its intended deployment regime. This benchmark is designed to mimic the common scenario in which many billions of web documents must be embedded once as a preprocessing stage for downstream organization and analysis.

\Cref{fig:throughput} summarizes end-to-end embedding throughput for \luxicalone{} and the transformer baselines. Even with GPU acceleration, the Qwen model lags behind \luxicalone{} by nearly two orders of magnitude. The much smaller MiniLM-based model closes part of this gap but still falls substantially short of \luxicalone{}, especially on CPU. These measurements confirm that the sparse-by-dense architecture and implementation choices in Luxical translate into practical speed improvements for large-scale corpus-processing workloads, not just improvements in FLOP counts on paper.

\subsection{Document-Half Matching}\label{subsec:half-matching}

To evaluate the utility of the embeddings produced by \luxicalone{}, we construct a web-document-based symmetrical retrieval task with known ground truth. To do this, we sample 50{,}000 documents from FineWeb and split each document into two contiguous halves, yielding 100{,}000 halves in total. For each original document, its two halves form a positive pair. We embed all halves with \luxicalone{} and treat each half in turn as a query. For a given query half, we compute cosine similarities to all 99{,}999 other halves and rank them. The matching half from the same source document defines the correct target; we record its rank and convert this to an error-at-$k$ curve as a function of the retrieval window size $k$. We compare against the same baseline models as in \Cref{subsec:throughput} as well as the following: \texttt{Arctic-2.0-M} (the teacher model for \luxicalone{} \citep{yu2024arcticembed20multilingualretrieval}), \texttt{LEAF-MT} (a model of the same size as \texttt{MiniLM-L6-v2} but trained using a similar knowledge-distillation objective \citep{vujanic2025leafknowledgedistillationtext}), and the MixedBreadAI-Large-v1 model (the teacher model for \texttt{LEAF-MT}, hereafter \texttt{Mxbai-L-v1} for brevity \citep{emb2024mxbai}).

\begin{figure}[htp]
  \centering
  \includegraphics[width=\linewidth]{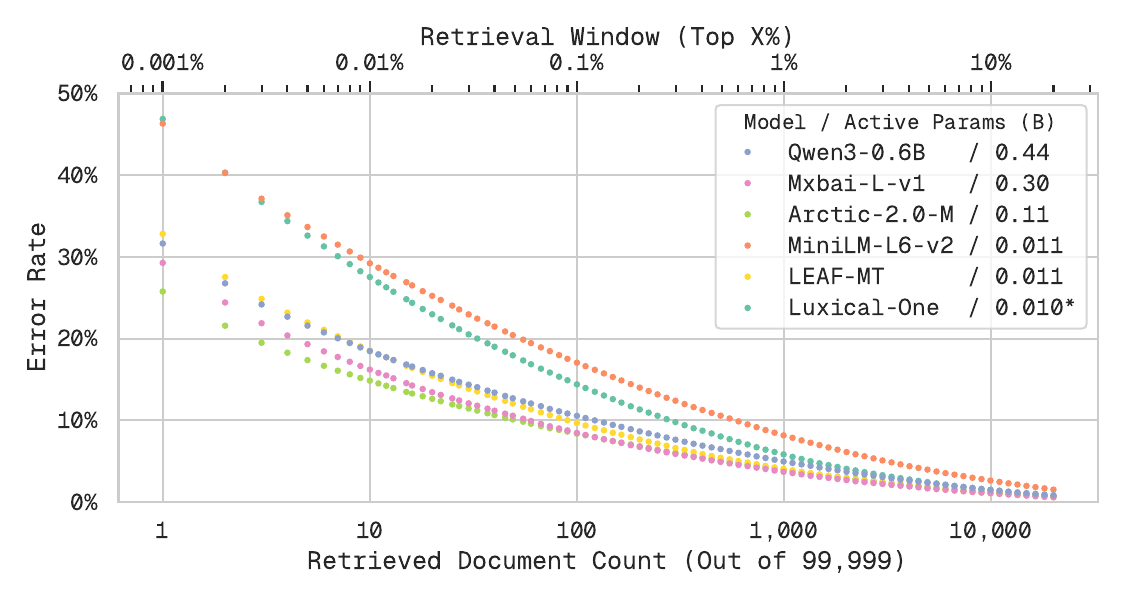}
  \caption{Document-half matching error rates as a function of retrieval window size for \luxicalone{} and transformer baselines on our document-half dataset.}
  \label{fig:doc-half-matching}
\end{figure}

\Cref{fig:doc-half-matching} reports error rates on the document-half matching task as a function of the retrieval window size. At strict top-1 retrieval, both \luxicalone{} and \texttt{MiniLM-L6-v2} trail far behind larger models like the Qwen model, though surprisingly \texttt{Arctic-2.0-M} does best despite its more modest active parameter count. As we enlarge the retrieval window, \luxicalone{} closes the gap and achieves substantially better error rates than the MiniLM-based baseline. These results show us that at coarse scales (which are potentially representative of many web-scale organization workloads like mining the top few percent of nearest neighbors to a target embedding vector), \luxicalone{}'s error curve approaches that of sophisticated transformer-based embedding models while maintaining dramatically higher throughput. These results suggest that Luxical can serve as an effective backbone for symmetric document--document similarity tasks in web-scale text organization pipelines. Its embedding geometry captures enough semantic structure to group related documents and support downstream classifiers, while its throughput makes it feasible to process very large corpora on commodity CPU hardware.

Another point worth taking away from this plot is that the distillation-based training objective is a powerful tool for training high-quality models. Even though \luxicalone{} embodies a more approximate function (e.g. lacking fine-grained positional information about the words of the input document), it is able to outperform the small transformer model by learning from a more powerful transformer model teacher. We see, too, that the \texttt{LEAF-MT} model strongly outperforms the same-architecture \texttt{MiniLM-L6-v2} model on this task, approaching the error rates of its teacher at coarse-grained retrieval windows. We speculate that both the distillation objective and the focus on full-document embedding to capture symmetrical similarity relationships were instrumental in making \texttt{LEAF-MT} perform so well on this task.

\subsection{Data Curation Application: Classifier-Based Filtering}\label{subsec:filtering}

\begin{figure}[htbp]
  \centering
  \begin{subfigure}[b]{\textwidth}
      \centering
      \includegraphics[width=\linewidth]{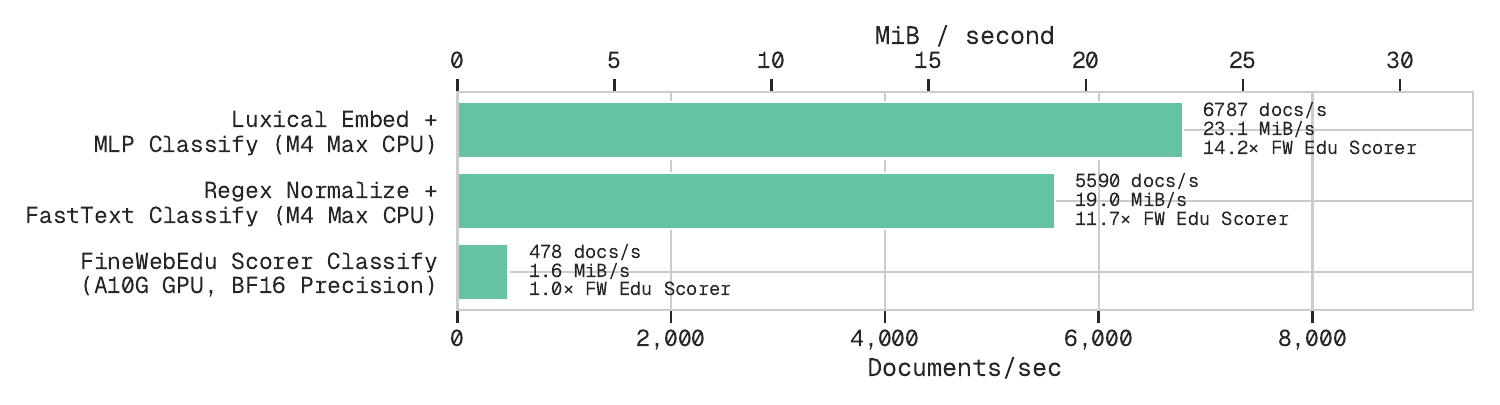}
      \caption{Classification throughput.}
      \label{fig:filtering-throughput}
  \end{subfigure}
  \hfill
  \begin{subfigure}[b]{\textwidth}
      \centering
      \includegraphics[width=\linewidth]{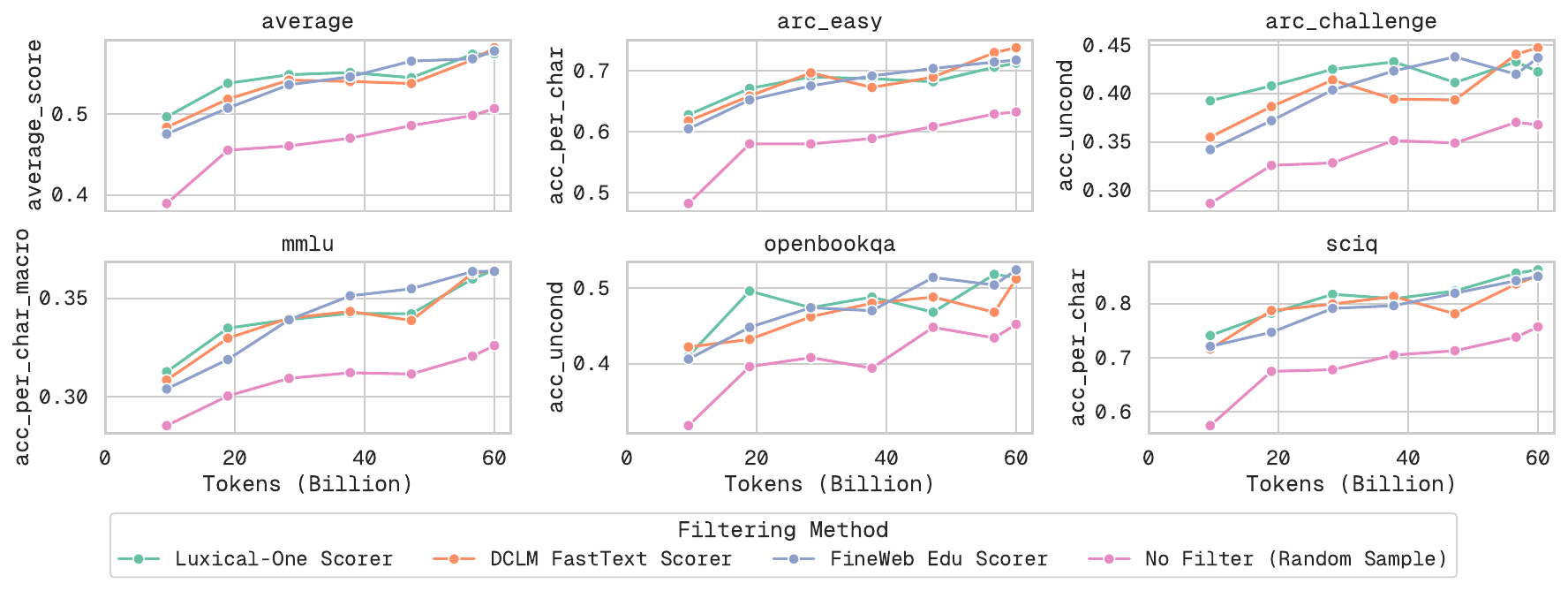}
      \caption{Downstream LM performance.}
      \label{fig:filtering-quality}
  \end{subfigure}
  \caption{Comparison of classifier-based filtering strategies. (a) Throughput of the scoring pipeline in documents per second. (b) Average zero-shot accuracy of a 3B-parameter language model trained on data curated by each scorer across 5 benchmarks.}
  \label{fig:filtering-results}
\end{figure}

To complement the document-half matching task in \Cref{subsec:half-matching}, which provides a controlled environment to probe the alignment between embedding geometry and semantic similarity, we additionally evaluate Luxical in a realistic end-to-end application: classifier-based data filtering for LLM training. Recent work has demonstrated that supervised text classifiers --- ranging from transformer-based encoders like the FineWeb-Edu scorer \citep{penedo2024fineweb} to lexical FastText classifiers like that used in DCLM \citep{li2024datacomp} --- can effectively identify high-quality subsets of web corpora, leading to improved downstream model performance. In this experiment, we compare \luxicalone{} against both the FineWeb-Edu and DCLM scoring models to assess whether our hybrid approach can match the utility of dense transformers while retaining the efficiency of lexical baselines.

We construct a filtering pipeline wherein a 600B token random subset of FineWeb is filtered down to a high-quality 60-billion-token subset (a 10\% selection rate).
We compare three scoring methods:
\begin{enumerate}
    \item \textbf{FineWeb-Edu Scorer:} We use the standard FineWeb-Edu classifier, which consists of a classification head trained on top of a BERT-like encoder network. The FineWeb-Edu classifier was trained using labels obtained by prompting Llama-3-70B-Instruct to score FineWeb documents for their educational quality.
    \item \textbf{DCLM FastText Scorer:} We use the standard DCLM scorer, a FastText classifier trained using samples from OpenHermes 2.5 \citep{OpenHermes2.5} and high-scoring posts from the r/ExplainLikeImFive subreddit as positives, and samples from a RefinedWeb \citep{penedo2023refinedwebdatasetfalconllm} reproduction as negatives.
    \item \textbf{Luxical-One MLP Scorer:} We train a lightweight Multi-Layer Perceptron (MLP) on top of frozen \luxicalone{} embeddings. The MLP consists of  two hidden layers with hidden dimensionality 256 and uses ReLU activations. We fit this MLP using a set of quality annotations similar to those used by the FineWeb-Edu and DCLM classifiers. We emphasize that the goal in this experiment is not to train a state-of-the-art text scorer, but to compare the performance characteristics of existing scorers to a \luxicalone{}-based scorer trained using similar data.
\end{enumerate}

To measure the impact of curation, we train a 3-billion-parameter dense transformer language model on each of the curated datasets. We use the AdamW optimizer \citep{adamw} in a warmup-stable-decay learning rate schedule reaching a peak learning rate of 7e-4 with a global batch size of 576. We evaluate the resulting models on a suite of filter-sensitive benchmarks, including ARC, MMLU, OpenBookQA, and SciQ \citep{clark2018arc, hendrycks2021mmlu, mihaylov2018openbookqa, SciQ}. To evaluate throughput, we time similar modeling pipelines in a controlled setting classifying 100,000 randomly-sampled FineWeb documents. 

\Cref{fig:filtering-results} summarizes the results of these experiments.
In terms of system throughput (\Cref{fig:filtering-throughput}), the \luxicalone{}-based scoring pipeline achieves speeds comparable to the FastText-based DCLM baseline (23.1MiB/s for \luxicalone{}-based pipeline, 19.0MiB/s for the FastText-based pipeline).
Crucially, both lexical approaches operate more than an order of magnitude faster than the transformer-based FineWeb-Edu scorer pipeline (which achieves a throughput of only 1.6MiB/s), even when the latter is GPU-accelerated. 
This confirms that Luxical successfully mitigates the computational bottlenecks associated with using BERT-style transformer models in this workflow.

Regarding quality (\Cref{fig:filtering-quality}), the language models trained on data curated by all three filtering methods achieve similar downstream accuracy, while the unfiltered baseline data yielded substantially lower performance.
These results suggest that for large-scale quality filtering, both FastText and \luxicalone{} offer a favorable trade-off on the speed--quality frontier, delivering utility comparable to that of heavy transformer encoders (e.g. inducing identical downstream MMLU accuracy scores of 36.4\%) at much greater throughput rates.
Additionally, since the MLP inference in this experiment accounts for less than 0.25\% of the total runtime of the \luxicalone{}-based classification pipeline, practitioners who wish to tweak and re-run scoring in an iterative manner can expect to accelerate their workflows beyond the speeds offered by FastText by decoupling an initial Luxical embedding step from a subsequent (and much faster) classification step.

\section{Discussion and Limitations}

Luxical is most appropriate when symmetric document--document similarity and coarse corpus-level organization are the primary goals, such as in semantic deduplication, clustering, and quality-based filtering at scale. In these regimes, a single fast embedding pass followed by geometric operations or lightweight classifiers can substantially reduce the cost of organizing large text corpora, and our experiments suggest that \luxicalone{} can replace heavier encoders without sacrificing coarse-grained quality when processing English web data. By contrast, Luxical is less suitable for settings that demand fine-grained ranking (e.g., high-precision search) or reasoning-heavy tasks, where larger transformer encoders remain the more reliable choice. For practitioners currently using MiniLM- or Qwen-style encoders for workloads like clustering, our results indicate that swapping in \luxicalone{} as the first-stage encoder is a plausible way to unlock the order-of-magnitude throughput gains observed in \Cref{fig:throughput} without re-architecting downstream components.

This report has several limitations. We study a single Luxical model, \luxicalone{}, trained on English FineWeb data and evaluated on one symmetric document similarity task and one language modeling data curation task, so additional work is needed to assess performance in other domains, languages, and task families. Our throughput comparison also focuses on a small set of hardware configurations and baseline models; a broader throughput and cost study across architectures and deployment settings is left for future work. Practitioners should therefore treat our results as an existence proof and reference implementation in one realistic setting, rather than a comprehensive comparison across all embedding choices.

Finally, there are natural extensions we do not explore here, including distilled variants, multilingual Luxical models, tighter integration with downstream corpus-management and curation systems, and connections to recent work on static embedding models in the sentence-transformers ecosystem (e.g. Static Embeddings \citep{Aarsen2025StaticEmbeddings} and model2vec \citep{minishlab2024model2vec}).

\subsection{Applications To Web-scale Text Organization}

In the setting described in \Cref{sec:intro,sec:background}, Luxical is intended to serve a single, reusable embedding stage in large-scale data pipelines. A typical deployment runs a Luxical model once over a corpus to produce dense vectors, then reuses those vectors across multiple downstream steps rather than invoking a model many times. Geometric methods such as clustering and nearest-neighbor search can operate directly on Luxical embeddings to construct domain-specific slices, perform distribution matching between training and evaluation sets, and implement semantic deduplication \citep{abbas2023semdedupdataefficientlearningwebscale} across billions of documents. The same embeddings can be fed into small classifiers to approximate expensive quality labels such as those used in FineWeb-Edu and DCLM-style FastText filters \citep{penedo2024fineweb,li2024datacomp}. In this regime, the cost of computing Luxical embeddings once is amortized across semantic deduplication, filtering, and ranking, making it practical to use richer model-based organization and curation signals at web scale on CPU. The document-half matching task in our experiments can be viewed as a controlled proxy for the symmetric similarity operations that underlie these deployment scenarios.

\section{Conclusion}

Luxical provides a practical point on the speed--quality frontier for web-scale text organization and filtering, sitting between purely lexical methods and full transformer encoders. By combining sparse TF--IDF features with a small neural network trained via Gram-matrix distillation, it delivers embeddings that are expressive enough for symmetric document similarity tasks while remaining inexpensive to compute on CPU. Our experiments with \luxicalone{} on a document-half-matching retrieval benchmark and a data curation classification task illustrate that this design can approach transformer-level coarse-grained quality at much higher throughput.

From a deployment perspective, Luxical is designed to be simple to integrate: models are small, CPU-friendly, and exposed through a straightforward API, making it easy to add a single embedding stage to existing corpus-processing pipelines and reuse the resulting vectors across multiple downstream tasks. The Luxical library is available as open-source software at \url{https://github.com/datologyai/luxical/}, and the \luxicalone{} model is hosted at \url{https://huggingface.co/DatologyAI/luxical-one}, along with example code that can serve as a starting point for reproducing and extending our results.

\section{Contributions and Acknowledgements}
\label{sec:contri}

\begin{tabularx}{\textwidth}{@{}p{0.21\textwidth}X@{}}

\textbf{Project Lead} & Luke Merrick \\[0.25em]
& \emph{conceptualized and implemented Luxical; trained \luxicalone{}; led evaluation} \\
\noalign{\vspace{0.75em}}

\textbf{Core Contributors} & Luke Merrick and Alex Fang \\[0.25em]
& \emph{conducted motivating research on fast lexical classification for data curation; conducted experiments evaluating \luxicalone{} on filter-based data curation} \\
\noalign{\vspace{0.75em}}

\textbf{Technical Contributors} & Aldo Carranza, Alvin Deng, Amro Abbas, Brett Larsen, Cody Blakeney, Darren Teh, David Schwab, Fan Pan, Haakon Mongstad, Haoli Yin, Jack Urbanek, Jason Lee, Jason Telanoff, Josh Wills, Kaleigh Mentzer, Parth Doshi, Paul Burstein, Pratyush Maini, Ricardo Monti, Rishabh Adiga, Scott Loftin, Siddharth Joshi, Spandan Das, Tony Jiang, Vineeth Dorna, and Zhengping Wang \\
\noalign{\vspace{0.25em}}
& \emph{DatologyAI technical staff; contributed the experimental pipelines used in \Cref{subsec:filtering}}\\
\noalign{\vspace{0.75em}}

\textbf{Not-So-Corporate \newline Leadership} & Bogdan Gaza, Ari Morcos, and Matthew Leavitt \\[0.25em]
\noalign{\vspace{0.75em}}

\textbf{Acknowledgements} &
Jacqueline Liu, Tiffanie Pham, and Sylvia Hoang for assembling the all-star cast that made this work possible. Liz Gatapia for the beautiful logo design. Jayla Lindsey for perpetuating the welcoming collaborative office environment that made this work possible. \\
\end{tabularx}

\bibliographystyle{plainnat}
\bibliography{luxical}

@article{DBLP:journals/corr/abs-1810.04805,
  author       = {Jacob Devlin and
                  Ming{-}Wei Chang and
                  Kenton Lee and
                  Kristina Toutanova},
  title        = {{BERT:} Pre-training of Deep Bidirectional Transformers for Language
                  Understanding},
  journal      = {CoRR},
  volume       = {abs/1810.04805},
  year         = {2018},
  url          = {http://arxiv.org/abs/1810.04805},
  archivePrefix = {arXiv},
  eprint       = {1810.04805}
}

@inproceedings{reimers-2019-sentence-bert,
  title     = {Sentence-BERT: Sentence Embeddings using Siamese {BERT}-Networks},
  author    = {Reimers, Nils and Gurevych, Iryna},
  booktitle = {Proceedings of the 2019 Conference on Empirical Methods in Natural Language Processing and the 9th International Joint Conference on Natural Language Processing (EMNLP-IJCNLP)},
  month     = nov,
  year      = {2019},
  address   = {Hong Kong, China},
  publisher = {Association for Computational Linguistics},
  url       = {https://aclanthology.org/D19-1410},
  doi       = {10.18653/v1/D19-1410},
  pages     = {3982--3992}
}

@inproceedings{muennighoff-etal-2023-mteb,
  title     = {{MTEB}: Massive Text Embedding Benchmark},
  author    = {Muennighoff, Niklas and Tunstall, Lewis and Shen, Long and Reimers, Nils and Mustafa, Badr and Harris, Michael and Tachbelie, Mita and Tiedemann, J{\"o}rg and Raja, Nouamane},
  booktitle = {Proceedings of the 17th Conference of the European Chapter of the Association for Computational Linguistics},
  month     = may,
  year      = {2023},
  address   = {Dubrovnik, Croatia},
  publisher = {Association for Computational Linguistics},
  url       = {https://aclanthology.org/2023.eacl-main.123},
  pages     = {1706--1726}
}

@article{DBLP:journals/corr/abs-1910.10683,
  author       = {Colin Raffel and
                  Noam Shazeer and
                  Adam Roberts and
                  Katherine Lee and
                  Sharan Narang and
                  Michael Matena and
                  Yanqi Zhou and
                  Wei Li and
                  Peter J. Liu},
  title        = {Exploring the Limits of Transfer Learning with a Unified Text-to-Text
                  Transformer},
  journal      = {CoRR},
  volume       = {abs/1910.10683},
  year         = {2019},
  url          = {http://arxiv.org/abs/1910.10683},
  archivePrefix = {arXiv},
  eprint       = {1910.10683}
}

@inproceedings{thibault-2021-splade,
  title     = {{SPLADE:} Sparse Lexical and Expansion Model for First Stage Ranking},
  author    = {Formal, Thibault and Piwowarski, Benjamin and Clinchant, St{\'e}phane},
  booktitle = {Proceedings of the 44th International {ACM} {SIGIR} Conference on Research and Development in Information Retrieval},
  year      = {2021},
  month     = jul,
  publisher = {Association for Computing Machinery},
  url       = {https://arxiv.org/abs/2107.05720},
  pages     = {1137--1146}
}

@article{penedo2024fineweb,
  title        = {The FineWeb Datasets: Decanting the Web for the Finest Text Data at Scale},
  author       = {Guilherme Penedo and Hynek Kydl{\'i}{\v{c}}ek and Loubna Ben Allal and Anton Lozhkov and Margaret Mitchell and Colin Raffel and Leandro von Werra and Thomas Wolf},
  journal      = {arXiv preprint arXiv:2406.17557},
  year         = {2024},
  url          = {https://arxiv.org/abs/2406.17557}
}

@article{li2024datacomp,
  title        = {{DataComp-LM}: In search of the next generation of training sets for language models},
  author       = {Li, Jeffrey and Fang, Alex and Smyrnis, Georgios and Ivgi, Maor and Jordan, Matt and Gadre, Samir and Bansal, Hritik and Guha, Etash and Keh, Sedrick and Arora, Kushal and Garg, Saurabh and Xin, Rui and Muennighoff, Niklas and Heckel, Reinhard and Mercat, Jean and Chen, Mayee and Gururangan, Suchin and Wortsman, Mitchell and Albalak, Alon and others},
  journal      = {arXiv preprint arXiv:2406.11794},
  year         = {2024},
  url          = {https://arxiv.org/abs/2406.11794}
}

@article{wang2022text,
  title   = {Text Embeddings by Weakly-Supervised Contrastive Pre-training},
  author  = {Wang, Liang and Yang, Nan and Cwikel, Howard and Salant, Shai and Almog, Oded and Halevy, Alon and Geva, Maor and Aharoni, Roee and Orlinsky, Arie},
  journal = {arXiv preprint arXiv:2212.03533},
  year    = {2022},
  url     = {https://arxiv.org/abs/2212.03533}
}

@inproceedings{joulin2017bag,
  title     = {Bag of Tricks for Efficient Text Classification},
  author    = {Joulin, Armand and Grave, Edouard and Bojanowski, Piotr and Mikolov, Tomas},
  booktitle = {Proceedings of the 15th Conference of the European Chapter of the Association for Computational Linguistics: Volume 2, Short Papers},
  year      = {2017},
  pages     = {427--431},
  publisher = {Association for Computational Linguistics}
}

@misc{penedo2023refinedwebdatasetfalconllm,
  title         = {The RefinedWeb Dataset for Falcon LLM: Outperforming Curated Corpora with Web Data, and Web Data Only},
  author        = {Guilherme Penedo and Quentin Malartic and Daniel Hesslow and Ruxandra Cojocaru and Alessandro Cappelli and Hamza Alobeidli and Baptiste Pannier and Ebtesam Almazrouei and Julien Launay},
  year          = {2023},
  eprint        = {2306.01116},
  archivePrefix = {arXiv},
  primaryClass  = {cs.CL},
  url           = {https://arxiv.org/abs/2306.01116}
}

@misc{soldaini2024dolmaopencorpustrillion,
  title         = {Dolma: an Open Corpus of Three Trillion Tokens for Language Model Pretraining Research},
  author        = {Luca Soldaini and Rodney Kinney and Akshita Bhagia and Dustin Schwenk and David Atkinson and Russell Authur and Ben Bogin and Khyathi Chandu and Jennifer Dumas and Yanai Elazar and Valentin Hofmann and Ananya Harsh Jha and Sachin Kumar and Li Lucy and Xinxi Lyu and Nathan Lambert and Ian Magnusson and Jacob Morrison and Niklas Muennighoff and Aakanksha Naik and Crystal Nam and Matthew E. Peters and Abhilasha Ravichander and Kyle Richardson and Zejiang Shen and Emma Strubell and Nishant Subramani and Oyvind Tafjord and Pete Walsh and Luke Zettlemoyer and Noah A. Smith and Hannaneh Hajishirzi and Iz Beltagy and Dirk Groeneveld and Jesse Dodge and Kyle Lo},
  year          = {2024},
  eprint        = {2402.00159},
  archivePrefix = {arXiv},
  primaryClass  = {cs.CL},
  url           = {https://arxiv.org/abs/2402.00159}
}

@misc{Aarsen2025StaticEmbeddings,
  author       = {Aarsen, Tom},
  title        = {Train 400x faster Static Embedding Models with Sentence Transformers},
  year         = {2025},
  month        = {Jan},
  publisher    = {{Hugging Face Blog}},
  howpublished = {\url{https://huggingface.co/blog/static-embeddings}}
}

@article{robertson2009probabilistic,
  author    = {Robertson, Stephen},
  title     = {The Probabilistic Relevance Framework: {BM25} and Beyond},
  journal   = {Foundations and Trends{\textregistered} in Information Retrieval},
  volume    = {3},
  number    = {4},
  pages     = {333--389},
  year      = {2009},
  publisher = {Now Publishers Inc.}
}

@misc{minishlab2024model2vec,
  author = {Tulkens, Stephan and {van Dongen}, Thomas},
  title = {Model2Vec: Fast State-of-the-Art Static Embeddings},
  year = {2024},
  url = {https://github.com/MinishLab/model2vec}
}

@inproceedings{lam2015numba,
  title={Numba: A llvm-based python jit compiler},
  author={Lam, Siu Kwan and Pitrou, Antoine and Seibert, Stanley},
  booktitle={Proceedings of the Second Workshop on the LLVM Compiler Infrastructure in HPC},
  pages={1--6},
  year={2015}
}

@misc{merrick2024arcticembedscalableefficientaccurate,
      title={Arctic-Embed: Scalable, Efficient, and Accurate Text Embedding Models}, 
      author={Luke Merrick and Danmei Xu and Gaurav Nuti and Daniel Campos},
      year={2024},
      eprint={2405.05374},
      archivePrefix={arXiv},
      primaryClass={cs.CL},
      url={https://arxiv.org/abs/2405.05374}, 
}

@article{spacesavingalgorithm,
author = {Zhao, Fuheng and Agrawal, Divyakant and Abbadi, Amr El and Metwally, Ahmed},
title = {SpaceSaving±: an optimal algorithm for frequency estimation and frequent items in the bounded-deletion model},
year = {2022},
issue_date = {February 2022},
publisher = {VLDB Endowment},
volume = {15},
number = {6},
issn = {2150-8097},
url = {https://doi.org/10.14778/3514061.3514068},
doi = {10.14778/3514061.3514068},
journal = {Proc. VLDB Endow.},
month = feb,
pages = {1215–1227},
numpages = {13}
}

@misc{abbas2023semdedupdataefficientlearningwebscale,
      title={SemDeDup: Data-efficient learning at web-scale through semantic deduplication}, 
      author={Amro Abbas and Kushal Tirumala and Dániel Simig and Surya Ganguli and Ari S. Morcos},
      year={2023},
      eprint={2303.09540},
      archivePrefix={arXiv},
      primaryClass={cs.LG},
      url={https://arxiv.org/abs/2303.09540}, 
}

@misc{sorscher2023neuralscalinglawsbeating,
      title={Beyond neural scaling laws: beating power law scaling via data pruning}, 
      author={Ben Sorscher and Robert Geirhos and Shashank Shekhar and Surya Ganguli and Ari S. Morcos},
      year={2023},
      eprint={2206.14486},
      archivePrefix={arXiv},
      primaryClass={cs.LG},
      url={https://arxiv.org/abs/2206.14486}, 
}

@misc{yu2024arcticembed20multilingualretrieval,
      title={Arctic-Embed 2.0: Multilingual Retrieval Without Compromise}, 
      author={Puxuan Yu and Luke Merrick and Gaurav Nuti and Daniel Campos},
      year={2024},
      eprint={2412.04506},
      archivePrefix={arXiv},
      primaryClass={cs.CL},
      url={https://arxiv.org/abs/2412.04506}, 
}

@misc{zhang2025qwen3embeddingadvancingtext,
      title={Qwen3 Embedding: Advancing Text Embedding and Reranking Through Foundation Models}, 
      author={Yanzhao Zhang and Mingxin Li and Dingkun Long and Xin Zhang and Huan Lin and Baosong Yang and Pengjun Xie and An Yang and Dayiheng Liu and Junyang Lin and Fei Huang and Jingren Zhou},
      year={2025},
      eprint={2506.05176},
      archivePrefix={arXiv},
      primaryClass={cs.CL},
      url={https://arxiv.org/abs/2506.05176}, 
}

@misc{sentence_transformers_all_minilm_l6_v2,
  title        = {Sentence-Transformers: all-MiniLM-L6-v2},
  author       = {{Sentence Transformers}},
  year         = {2021},
  howpublished = {Hugging Face Model Hub},
  url          = {https://huggingface.co/sentence-transformers/all-MiniLM-L6-v2}
}

@misc{vujanic2025leafknowledgedistillationtext,
      title={LEAF: Knowledge Distillation of Text Embedding Models with Teacher-Aligned Representations}, 
      author={Robin Vujanic and Thomas Rueckstiess},
      year={2025},
      eprint={2509.12539},
      archivePrefix={arXiv},
      primaryClass={cs.IR},
      url={https://arxiv.org/abs/2509.12539}, 
}

@online{emb2024mxbai,
  title  = {Open Source Strikes Bread - New Fluffy Embeddings Model},
  author = {Sean Lee and Aamir Shakir and Darius Koenig and Julius Lipp},
  year   = {2024},
  url    = {https://www.mixedbread.ai/blog/mxbai-embed-large-v1}
}

@misc{clark2018arc,
      title={Think you have Solved Question Answering? Try ARC, the AI2 Reasoning Challenge}, 
      author={Peter Clark and Isaac Cowhey and Oren Etzioni and Tushar Khot and Ashish Sabharwal and Carissa Schoenick and Oyvind Tafjord},
      year={2018},
      eprint={1803.05457},
      archivePrefix={arXiv},
      primaryClass={cs.AI},
      url={https://arxiv.org/abs/1803.05457}, 
}

@misc{hendrycks2021mmlu,
      title={Measuring Massive Multitask Language Understanding}, 
      author={Dan Hendrycks and Collin Burns and Steven Basart and Andy Zou and Mantas Mazeika and Dawn Song and Jacob Steinhardt},
      year={2021},
      eprint={2009.03300},
      archivePrefix={arXiv},
      primaryClass={cs.CY},
      url={https://arxiv.org/abs/2009.03300}, 
}

@misc{mihaylov2018openbookqa,
      title={Can a Suit of Armor Conduct Electricity? A New Dataset for Open Book Question Answering}, 
      author={Todor Mihaylov and Peter Clark and Tushar Khot and Ashish Sabharwal},
      year={2018},
      eprint={1809.02789},
      archivePrefix={arXiv},
      primaryClass={cs.CL},
      url={https://arxiv.org/abs/1809.02789}, 
}

@inproceedings{SciQ,
    title={Crowdsourcing Multiple Choice Science Questions},
    author={Johannes Welbl, Nelson F. Liu, Matt Gardner},
    year={2017},
    journal={arXiv:1707.06209v1}
}

@misc{adamw,
      title={Decoupled Weight Decay Regularization}, 
      author={Ilya Loshchilov and Frank Hutter},
      year={2019},
      eprint={1711.05101},
      archivePrefix={arXiv},
      primaryClass={cs.LG},
      url={https://arxiv.org/abs/1711.05101}, 
}

@misc{OpenHermes2.5,
  title = {OpenHermes 2.5: An Open Dataset of Synthetic Data for Generalist LLM Assistants},
  author = {Teknium},
  year = {2023},
  publisher = {HuggingFace},
  url = {https://huggingface.co/datasets/teknium/OpenHermes-2.5}
}

\end{document}